\documentclass[11pt]{article}
\usepackage{amsmath}
\usepackage{times}
\usepackage{graphicx}
\usepackage{color}
\usepackage{multirow}
\usepackage[authoryear]{natbib}
\usepackage{rotating}
\usepackage{bbm}
\usepackage{latexsym}
\usepackage{booktabs}

\usepackage[utf8]{inputenc} 
\usepackage[T1]{fontenc}    
\usepackage{hyperref}       
\usepackage{url}            
\usepackage{booktabs}       
\usepackage{amsfonts}       
\usepackage{nicefrac}       
\usepackage{microtype}      
\usepackage{xcolor}         

\usepackage{graphicx}

\title{Scaled Dot-Product Attention implements projection of inputs onto a common surface}

\author{%
Terence D.~Sanger\\
  Department of Electrical Engineering and Computer Science\\
University of California, Irvine\\
Irvine, California 92697 \\
  \texttt{terry@~sangerlab.net} \\
}

\begin{document}

\maketitle

\begin{abstract}
Scaled dot-product attention (SDPA) is a fundamental component responsible for the success of large-language models and other nonlinear signal processing applications. The rationale for SDPA has been based upon "query, key, value" concepts borrowed from database theory, but these concepts are difficult to reconcile with standard methods in mathematical signal processing. We show that SDPA can be rewritten in a different but mathematically equivalent form as a projection of the input vectors onto a common surface determined by the inputs themselves. Therefore SDPA discovers nonlinear dependencies in the input that are time-dependent and context-dependent. The rewritten form of SDPA permits increased speed of both feedforward and learning algorithms, but more importantly suggests potential extensions. In the context of language, we re-interpret the role of SDPA as finding a time-dependent contextual meaning determined  by the surface on which the set of input vectors lies. Input token embeddings are then modified by the local context surface. This interpretation differs substantially from the concept of "self-attention", and provides a strong justification for the use of SDPA for time-series data with time-varying local nonlinear dependencies.
\end{abstract}

\section{Introduction}

Scaled dot-product attention (SDPA) is an essential component of many successful machine learning models\citep{niu2021review}.  This mechanism was originally derived from database indexing concepts, and relies upon "query", "key", and "value" vectors that identify long-range interactions between inputs at different times\citep{vaswani2017attention,wolf-etal-2020-transformers,vig2019analyzing,clark-etal-2019-bert}.  Unfortunately, this model for the function of SDPA does not allow mathematical analysis of the function of the algorithm, and it does not lead immediately to new directions for further development\citep{ahmed2023transformers,anwar2024transformers,yildiz2022multivariate,yang2020complex,sohn2024implementing,cui2024phase,phuong2022formal,ferrando2024primer}.  A mathematical description of SDPA would allow linkage to concepts in signal processing and dynamic systems theory that may significantly expand the impact of this algorithm.  Furthermore, it is important to understand whether SDPA is in fact a new computational mechanism with significant added power compared to standard signal processing methods, or whether it is a new application of an existing and known algorithm that might already have extensive analysis.  

An early example of the value of this type of mathematical analysis was the application of linear algebra to the Perceptron model, originally developed and inspired by neuroscience.  The use of linear algebra allowed significant advances, including the eventual development of the backpropagation weight-fitting algorithm.  We seek a similarly useful mathematical basis for SDPA.

Here, we will show that SDPA is equivalent to a projection operation.  Specifically, SDPA projects input samples onto a surface defined by nearby input samples.  When the samples come from a time-series, this means that SDPA finds a time-varying surface defined by linear relations between the incoming data points.  The methods we use are essentially a rewriting of the SDPA equations, using distance between samples rather than the dot product.  

\section{Methods}

Consider three matrices $q = Q W_q$, $k =  K W_k$, and $v = V W_v $ where $Q$, $K$, and $V$ are data encodings with rows $Q(t), K(t), V(t)$, and $W_q$,  $W_k$, and $W_v$ are learned linear maps.  The SDPA layer implements the function
\begin{equation}
	y = {\rm softmax}\left( \frac{qk^T}{\sqrt{d}} \right) v 
	\label{eq-sdpa}
\end{equation}
where $d$ is the dimensionality of the encoding (embedding) space.  We will ignore $d$ in the analysis, since it can easily be absorbed in $W_q$ and $W_k$.  For cross-attention we usually have $k=v$, and for self attention we further have $q=k=v$.  We will show the derivation for cross-attention, since self-attention is a special case.  We can write:
\begin{equation}
	y = {\rm softmax}\left( qk^T \right) k 
\end{equation}
realizing that if $W_k \neq W_v$ then we can make the layer equivalent to the case $W_k = W_v$ by adding a linear transform at the output.

Layer normalization is an essential component of the Transformer architecture\citep{xiong2020layer}, and we can therefore assume that $\|q(i) \| = \|k(j) \| = 1$ for all time indices $i,j$, where $q(i)$ and $k(j)$ are row vectors.  We thus have:
\begin{eqnarray}
	\| q(i) - k(j) \| & = & \| q(i) \|^2 + \| k(j) \|^2 - 2 q(i) k(j)^T \\
	& = & 2 - 2 q(i) k(j)^T\\
	q(i) k(j)^T &=& 1 - \| q(i) - k(j) \|/2
\end{eqnarray}
Define the distance matrix $d_{ij} =  \| q(i) - k(j) \|$, and define the matrix $z = {\rm softmax}(qk^T)$ so that applying the softmax() operator along the $j$ dimension gives
\begin{eqnarray}
	z_{ij} & = & \frac{e^{1-d_{ij}/2} }{\sum_j e^{1-d_{ij}/2}} \\
	& = &  \frac{e^{-d_{ij}/2} }{\sum_j e^{-d_{ij}/2}} \\
	& = &  \frac{1}{C_i}e^{-\| q(i) - k(j) \|/2}  \\
\end{eqnarray}
for constants $C_i$ that cause the row sums to be 1.  This form should be familiar as the probability of observing the vector $k(j)$ under a Gaussian distribution with mean $q(i)$ and variance 1.  We extend the algorithm slightly by allowing tunable variance $\sigma^2$:
\begin{eqnarray}
	z_{ij} & = &  \frac{1}{C_i}e^{-\| q(i) - k(j) \|/2\sigma^2}  \\
\end{eqnarray}

Now if $y_i$ is the $i$'th row of the output matrix and $z_i$ is the $i$'th row of the weight matrix, then
\begin{eqnarray}
	y_i & = & z_i k \\
\end{eqnarray}
In other words, $z_{ij}$ is the weighting for the contribution of input vector $k_j$ to the vector output $y_i$.   The rewritten "projection SDPA" algorithm is thus
\begin{equation}
	y_{im} = \sum_j  \frac{1}{C_i}e^{-\| q(i) - k(j) \|/2\sigma^2} k_{jm}
	\label{sdpa-proj}
\end{equation}

\begin{figure}
  \centering
 \includegraphics[width=5in]{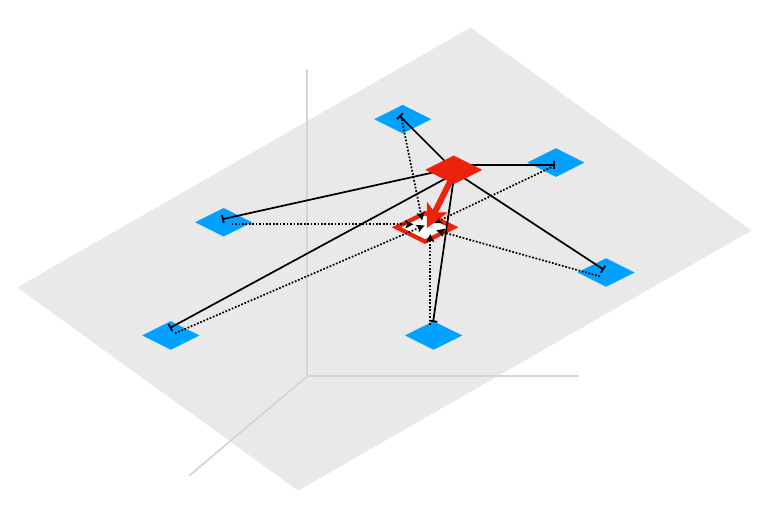}
  \caption{Illustration of projection SDPA. Blue squares are nearby/prior input vectors, solid red square is the current input vector.  The output of the layer is shown by the open red square which is a linear combination of the nearby vectors.  The combination weights (dotted lines) are proportional to the Gaussian-weighted distance (solid lines) from the current vector to the nearby/prior vectors.  The overall effect of the layer is to project each input (solid red square) onto an output vector (open red square) given by a linear combination of other inputs (blue squares).  }
  \label{fig1}
\end{figure}

Therefore we have shown that SDPA projects each query input $q(i)$ onto a linear combination of the key vectors $k(j)$ with projection weights determined by a Gaussian probability distribution.  For self-attention when $q=k$, each input sample is projected onto the space of all the other samples, effectively "flattening" the set of inputs into a common linear subspace (see figure \ref{fig1}).

Note that if causal masking is used to prevent interactions between samples and future samples, this is implemented by setting all above-diagonal elements of the matrix 
\begin{equation}
	e^{-d_{ij}/2}
\end{equation}
to be zero, prior to row normalization
\begin{equation}
	 \frac{e^{-d_{ij}/2} }{\sum_j e^{-d_{ij}/2}}
\end{equation}
so that masking effectively projects each sample onto the space of linear combinations of prior samples.  Multi-head attention is implemented in the standard way, by slicing the input vectors and using separate projection operators for each slice, subsequently recombining using a linear layer.

\section{Results}

Since projection SDPA is simply a rewriting of the SDPA equation, we do not expect it to have any additional computational power compared to standard SDPA.  However, it is possible that performance under a backpropagation learning algorithm would be different.  It is important to realize that the projection operator itself has no learnable parameters (unless $\sigma$ is learned), since all learnable parameters are in the input matrices $W_q, W_k, W_v$, similarly to standard SDPA.

\begin{figure}
  \centering
a.
\includegraphics[width=2.5in]{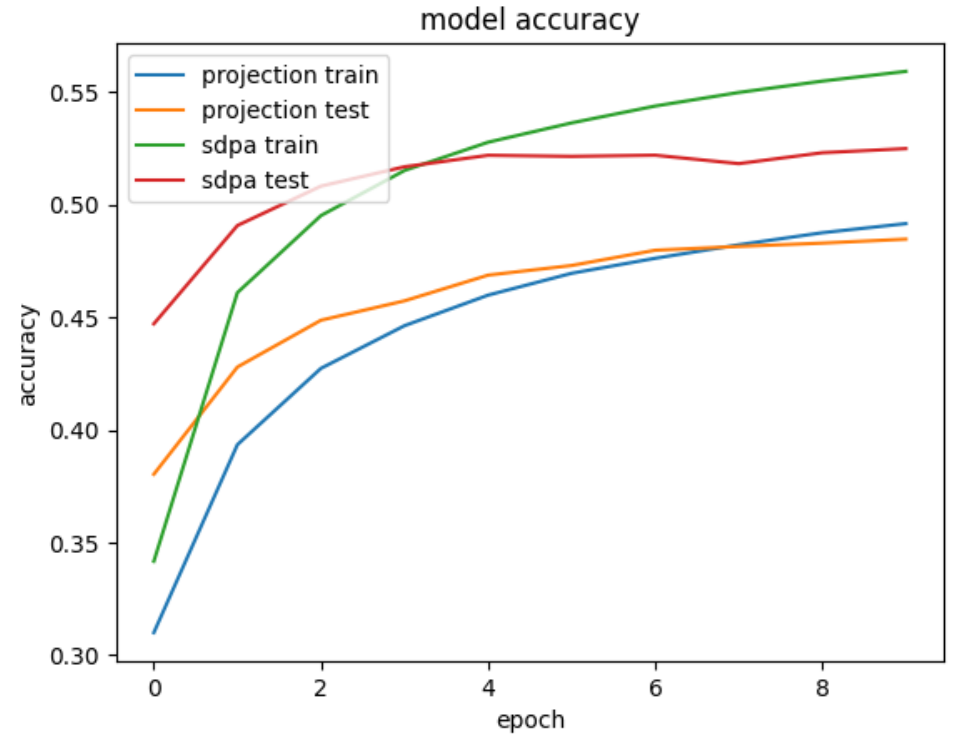}
b.
\includegraphics[width=2.5in]{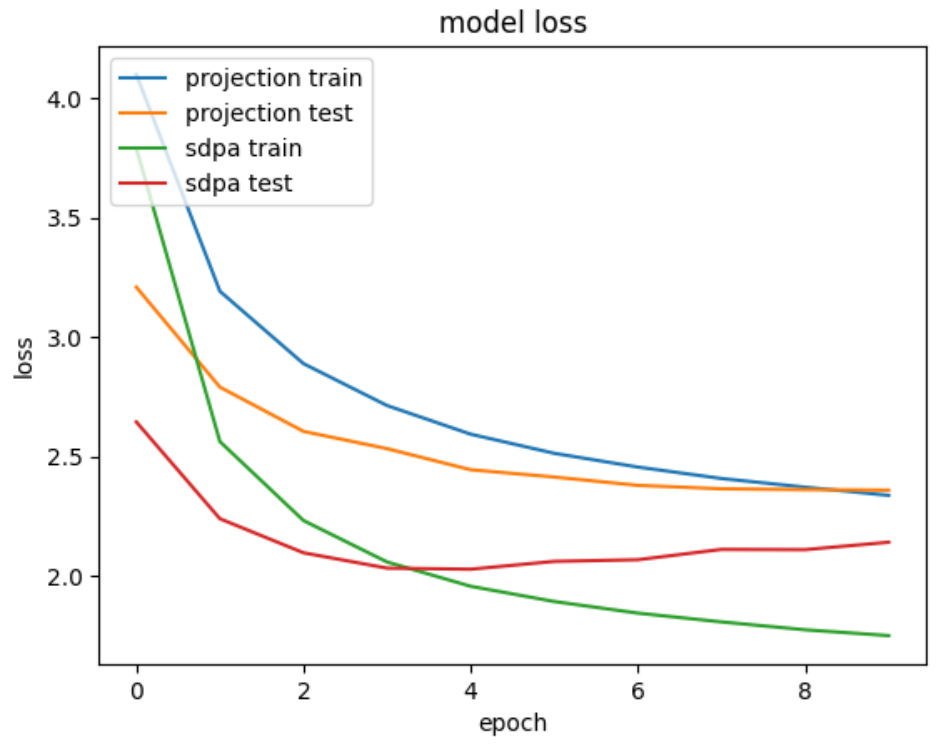}

\caption{a. Accuracy (fraction of correct word matches) on test set and validation set during training for standard and projection SDPA models during 10 epochs of training.  b. Cross-entropy loss on training set and validation set.}
\label{fig2}
\end{figure}

To test this, we modified publicly-available standard code for a Transformer model by substituting equation \ref{sdpa-proj} for equation \ref{eq-sdpa}.  (Code is available in the supplementary material.)  Standard SDPA and projection SDPA were tested on a Spanish-to-English translation task.  Source data are derived from the Tatoeba.org archive, with 118,000 phrase/sentence pairs of increasing length.  70\% of the pairs were used for training, 15\% for validation, and 15\% for testing.  The encoding used a vocabulary size of 15000, sequence length of 10 with zero-padding, and the standard transformer architecture as in \citet{vaswani2017attention} with 8 multi-heads for each model.  $\sigma$ for the Gaussian weighting was 0.01 for the self-attention and 0.05 for the cross-attention modules.  Causal masking was used on the decoder input stage but not on the encoder stage.  Padding masking was not used.  Training of each model occurred over 10 epochs.  Results for cross-entropy loss and target accuracy (fraction of words correctly identified) are shown during learning in figure \ref{fig2}.    Figure \ref{fig3} shows a selection of randomly-chosen prompts and output sentences.

Training (10 epochs) required 174 seconds for the standard SDPA, and 129 seconds for the projection SDPA, running on an A100 GPU virtual machine under Google Colab(tm).  The accuracy results for the standard SDPA were slightly better than for projection SDPA, and learning appeared to converge faster, suggesting that the training for standard SDPA is more efficient despite the slower speed.  

\begin{figure}
\begin{quotation}
Please don't cry again.
[start] por favor no llorar [end]

What do we owe you?
[start] qué le debo [end]

I decided to continue doing that.
[start] decidió seguir haciendo eso [end]

I don't want to hurt you, Tom.
[start] no quiero herido [end]

Let me go alone.
[start] déjame ir sola [end]

She kept crying all night long.
[start] ella siguió llorando todo el tiempo [end]
\end{quotation}
\caption{Sample input prompts (English) and network outputs (Spanish) after training for Transformer model using projection SDPA layers.}
\label{fig3}
\end{figure}

\section{Discussion}

We have proposed a re-interpretation of the mechanism of Scaled Dot-product Attention (SDPA) in terms of projection of input samples onto a surface determined by nearby input samples.  While this interpretation is easily representable in standard mathematical concepts, it nevertheless may be a unique and rarely, if ever, used technique in signal processing prior to the advent of Transformers.  One reason for this is that projection SDPA requires maintaining a history of recent samples, and this differs from the standard signal processing and dynamic systems approach in which information from all prior samples is maintained in a state vector.  Of course, an augmented state vector could be created to hold recent samples, but this is not a common approach.  Furthermore, the projection operation is cubic in the input values and thus not representable as a linear system.  Therefore we believe that SDPA is indeed a novel signal processing method with significant additional power compared to standard recursive processing methods.

The relationship to the original components of the SDPA and transformer models is informative.  The softmax operation within standard SDPA implements a Gaussian probability density within projection SDPA.  The layer normalization within standard SDPA is essential for creating the equivalence between dot-product and distance operators in projection SDPA.  The layer bypass operation (in which inputs to an SDPA layer are added to the outputs) seems to be essential to avoid collapse of the low-dimensional surface onto progressively lower and lower dimensional surfaces until all data are projected to a single point.  The linear input matrices $W_q, W_k, W_v$ scale the dimensions of the embedding space and thus control the dimensions with greatest contribution to the distances $\| q-k \|$ that determine the mixing weights.

The self-attention formulation ($k=v=q$) corresponds to projection of each input sample onto the space of nearby samples, thus performing a local "flattening" operation.  In other words, the input sequence is constrained to lie closer to a low-dimensional surface within the embedding space.  This operation may reduce noise.  In the context of language, it can be interpreted as constraining the interpretation of words to be consistent with the context defined by recent words.  This is less an "attentional" mechanism and more of a "consistency" mechanism.  The projection reduces the effective dimensionality of the input data and thus may have similar benefits to other dimensionality reduction methods.  

The cross-attention formulation ($k=v \neq q$) corresponds to projection of input samples onto a space defined by a separate set of context measurements.  In the case of language translation, this means projecting the translation onto the space defined by the untranslated prompt sentence.  In the example above, this means that the enlarging Spanish sentence is projected both onto itself (maintaining grammatical consistency, fluency, and  meaning) and also onto the English prompt sentence (maintaining consistency with the meaning of the untranslated original).

As currently formulated, SDPA uses a finite set of past samples and thus fits in the class of Finite Impulse Response (FIR) filters.  An Infinite Impulse Response (IIR) version would require a time weighting that decays backward in time, and this might have some important uses for Robotics, Signal processing, and interpretation of very long texts.  The current formulation, as with almost all Transformer architectures, operates only in discrete time.  Continuous-time versions are possible, although computationally expensive due to the need to compute integrals of the form
\begin{equation}
	d(t,s) \sim \int q(t-\tau_1) k^T(s-\tau_2) d\tau_1 d\tau_2
\end{equation}
for all values of $s$ and $t$.  

\section{Conclusion}

We have discussed  the mathematical derivation of projection SDPA and shown that it exhibits similar computational and learning properties to standard SDPA.  Further work will be needed to optimize this algorithm, explore $\sigma$ and possibly other parameters, and explore potential extensions to this method.  We hope that by providing an alternative mathematically-based explanation of the SDPA layer that further development, understanding, and improvement will be possible in the future.

\bibliographystyle{apalike}
\bibliography{Transformers}

@article{vaswani2017attention,
  title={Attention is all you need},
  author={Ashish Vaswani and Noam Shazeer and Niki Parmar and Jakob Uszkoreit and Llion Jones and Aidan N. Gomez and Lukasz Kaiser and  Illia Polosukhin},
  journal={Advances in Neural Information Processing Systems},
  year={2017},
  pages={5998--6008}
}

@inproceedings{wolf-etal-2020-transformers,
    title = "Transformers: State-of-the-Art Natural Language Processing",
    author = "Wolf, Thomas  and
      Debut, Lysandre  and
      Sanh, Victor  and
      Chaumond, Julien  and
      Delangue, Clement  and
      Moi, Anthony  and
      Cistac, Pierric  and
      Rault, Tim  and
      Louf, Remi  and
      Funtowicz, Morgan  and
      Davison, Joe  and
      Shleifer, Sam  and
      von Platen, Patrick  and
      Ma, Clara  and
      Jernite, Yacine  and
      Plu, Julien  and
      Xu, Canwen  and
      Le Scao, Teven  and
      Gugger, Sylvain  and
      Drame, Mariama  and
      Lhoest, Quentin  and
      Rush, Alexander",
    editor = "Liu, Qun  and
      Schlangen, David",
    booktitle = "Proceedings of the 2020 Conference on Empirical Methods in Natural Language Processing: System Demonstrations",
    month = oct,
    year = "2020",
    address = "Online",
    publisher = "Association for Computational Linguistics",
    url = "https://aclanthology.org/2020.emnlp-demos.6/",
    doi = "10.18653/v1/2020.emnlp-demos.6",
    pages = "38--45"
}

@inproceedings{xiong2020layer,
  title={On layer normalization in the transformer architecture},
  author={Xiong, Ruibin and Yang, Yunchang and He, Di and Zheng, Kai and Zheng, Shuxin and Xing, Chen and Zhang, Huishuai and Lan, Yanyan and Wang, Liwei and Liu, Tieyan},
  booktitle={International Conference on Machine Learning},
  pages={10524--10533},
  year={2020},
  organization={PMLR}
}

@inproceedings{clark-etal-2019-bert,
    title = "What Does {BERT} Look at? An Analysis of {BERT}`s Attention",
    author = "Clark, Kevin  and
      Khandelwal, Urvashi  and
      Levy, Omer  and
      Manning, Christopher D.",
    editor = "Linzen, Tal  and
      Chrupa{\l}a, Grzegorz  and
      Belinkov, Yonatan  and
      Hupkes, Dieuwke",
    booktitle = "Proceedings of the 2019 ACL Workshop BlackboxNLP: Analyzing and Interpreting Neural Networks for NLP",
    month = aug,
    year = "2019",
    address = "Florence, Italy",
    publisher = "Association for Computational Linguistics",
    url = "https://aclanthology.org/W19-4828/",
    doi = "10.18653/v1/W19-4828",
    pages = "276--286"
}

@article{phuong2022formal,
  title={Formal algorithms for transformers},
  author={Phuong, Mary and Hutter, Marcus},
  journal={arXiv preprint arXiv:2207.09238},
  year={2022}
}

@article{ferrando2024primer,
  title={A primer on the inner workings of transformer-based language models},
  author={Ferrando, Javier and Sarti, Gabriele and Bisazza, Arianna and Costa-juss{\`a}, Marta R},
  journal={arXiv preprint arXiv:2405.00208},
  year={2024}
}

@article{ahmed2023transformers,
  title={Transformers in time-series analysis: A tutorial},
  author={Ahmed, Sabeen and Nielsen, Ian E and Tripathi, Aakash and Siddiqui, Shamoon and Ramachandran, Ravi P and Rasool, Ghulam},
  journal={Circuits, Systems, and Signal Processing},
  volume={42},
  number={12},
  pages={7433--7466},
  year={2023},
  publisher={Springer}
}

@article{anwar2024transformers,
  title={Transformers in biosignal analysis: A review},
  author={Anwar, Ayman and Khalifa, Yassin and Coyle, James L and Sejdic, Ervin},
  journal={Information Fusion},
  pages={102697},
  year={2024},
  publisher={Elsevier}
}

@article{yildiz2022multivariate,
  title={Multivariate time series imputation with transformers},
  author={Y{\i}ld{\i}z, A Yark{\i}n and Ko{\c{c}}, Emirhan and Ko{\c{c}}, Aykut},
  journal={IEEE Signal Processing Letters},
  volume={29},
  pages={2517--2521},
  year={2022},
  publisher={IEEE}
}

@inproceedings{yang2020complex,
  title={Complex transformer: A framework for modeling complex-valued sequence},
  author={Yang, Muqiao and Ma, Martin Q and Li, Dongyu and Tsai, Yao-Hung Hubert and Salakhutdinov, Ruslan},
  booktitle={ICASSP 2020-2020 IEEE International Conference on Acoustics, Speech and Signal Processing (ICASSP)},
  pages={4232--4236},
  year={2020},
  organization={IEEE}
}

@article{sohn2024implementing,
  title={Implementing and Optimizing the Scaled Dot-Product Attention on Streaming Dataflow},
  author={Sohn, Gina and Zhang, Nathan and Olukotun, Kunle},
  journal={arXiv preprint arXiv:2404.16629},
  year={2024}
}

@article{niu2021review,
  title={A review on the attention mechanism of deep learning},
  author={Niu, Zhaoyang and Zhong, Guoqiang and Yu, Hui},
  journal={Neurocomputing},
  volume={452},
  pages={48--62},
  year={2021},
  publisher={Elsevier}
}

@article{cui2024phase,
  title={A phase transition between positional and semantic learning in a solvable model of dot-product attention},
  author={Cui, Hugo and Behrens, Freya and Krzakala, Florent and Zdeborov{\'a}, Lenka},
  journal={arXiv preprint arXiv:2402.03902},
  year={2024}
}

@article{vig2019analyzing,
  title={Analyzing the structure of attention in a transformer language model},
  author={Vig, Jesse and Belinkov, Yonatan},
  journal={arXiv preprint arXiv:1906.04284},
  year={2019}
}

\end{document}